\documentclass[lettersize,journal]{IEEEtran}
\usepackage{subcaption}
\usepackage{amsmath,amsfonts}
\usepackage{algorithmic}
\usepackage{array}
\usepackage{textcomp}
\usepackage{stfloats}
\usepackage{url}
\usepackage{verbatim}
\usepackage{graphicx}
\usepackage{enumitem}
\usepackage{color}
\usepackage{makecell}
\usepackage{ragged2e} 
\usepackage{booktabs,makecell, multirow, tabularx}

\usepackage[linesnumbered,boxed,ruled,commentsnumbered]{algorithm2e}

\usepackage{cite}
\begin{document}

\title{TMLC-Net: Transferable Meta Label Correction for Noisy Label Learning}

\author{Mengyang Li
\thanks{M. Li is with School of Artificial Intelligence, Tianjin Normal University, Tianjin 300074, China (e-mail: limengyang@tjnu.edu.cn).}
\thanks{Manuscript received July 6, 2024.}}

\markboth{Journal of \LaTeX\ Class Files,~Vol.~xx, No.~xx, xxxx~xxxx}%
{Shell \MakeLowercase{\textit{et al.}}: A Sample Article Using IEEEtran.cls for IEEE Journals}

\maketitle

\begin{abstract}
The prevalence of noisy labels in real-world datasets poses a significant impediment to the effective deployment of deep learning models. While meta-learning strategies have emerged as a promising approach for addressing this challenge, existing methods often suffer from limited transferability and task-specific designs.  This paper introduces TMLC-Net, a novel Transferable Meta-Learner for Correcting Noisy Labels, designed to overcome these limitations. TMLC-Net learns a general-purpose label correction strategy that can be readily applied across diverse datasets and model architectures without requiring extensive retraining or fine-tuning. Our approach integrates three core components: (1) Normalized Noise Perception, which captures and normalizes training dynamics to handle distribution shifts; (2) Time-Series Encoding, which models the temporal evolution of sample statistics using a recurrent neural network; and (3) Subclass Decoding, which predicts a corrected label distribution based on the learned representations.  We conduct extensive experiments on benchmark datasets with various noise types and levels, demonstrating that TMLC-Net consistently outperforms state-of-the-art methods in terms of both accuracy and robustness to label noise. Furthermore, we analyze the transferability of TMLC-Net, showcasing its adaptability to new datasets and noise conditions, and establishing its potential as a broadly applicable solution for robust deep learning in noisy environments.
\end{abstract}

\section{Introduction}
\label{sec:introduction}

Deep learning has achieved remarkable success across a wide range of domains, including computer vision, natural language processing, and speech recognition \cite{krizhevsky2012imagenet, he2016deep, devlin2018bert, hinton2012deep}. This success is largely driven by the availability of massive datasets and the development of powerful neural network architectures. However, a critical, often overlooked assumption underlying the effectiveness of these models is the accuracy of the labels in the training data. In real-world scenarios, this assumption is frequently violated due to factors such as human annotation errors, imperfect data collection processes, and the inherent subjectivity of certain labeling tasks \cite{frenay2013classification, northcutt2021pervasive}. The presence of noisy labels can significantly compromise the performance, reliability, and generalization capability of deep learning models \cite{zhang2021understanding, algan2021image}, leading to overfitting, bias amplification, and reduced robustness to adversarial attacks \cite{arpit2017closer, buolamwini2018gender, goodfellow2014explaining, szegedy2013intriguing}.

Addressing the challenge of noisy labels has led to a substantial body of research, broadly categorized into loss correction, sample selection, and label correction techniques \cite{ghosh2017robust, hendrycks2018deep, patrini2017making, jiang2018mentornet, han2018co, reed2014training, tanaka2018joint}. Loss correction methods modify the loss function to be less sensitive to noisy labels, using robust loss functions or loss adjustment strategies. Sample selection methods aim to identify and remove or down-weight noisy samples. Label correction methods attempt to directly modify the incorrect labels. Despite the progress, many existing methods rely on strong assumptions about the underlying noise distribution (e.g., class-conditional or uniform noise), exhibit limited adaptability to new situations, and incur high computational costs. Meta-learning has emerged as a promising direction \cite{ren2018meta, li2019learning, yu2019does}, but existing meta-label correction methods often fall short in terms of transferability, hindering their practical utility and increasing computational overhead due to bi-level optimization.

To overcome these limitations, we introduce TMLC-Net, a Transferable Meta-Learner for Correcting Noisy Labels. TMLC-Net is explicitly designed to learn a general-purpose label correction strategy that can be effectively transferred across diverse datasets and model architectures without requiring extensive retraining or fine-tuning. This transferability is a key differentiator from existing meta-label correction approaches. TMLC-Net achieves this through a combination of: (1) Normalized Noise Perception, capturing and normalizing training dynamics to handle distribution shifts; (2) Time-Series Encoding, modeling the temporal evolution of sample statistics using an LSTM; and (3) Subclass Decoding, predicting a corrected label distribution for more informed correction.

The proposed TMLC-Net offers several advantages: it addresses the critical limitation of transferability in existing methods, enhances robustness to distribution shifts via normalized noise perception, captures the dynamic learning process through time-series encoding, and provides a more informed and less brittle label correction via subclass decoding.

Our contributions are summarized as follows:

\begin{itemize}
    \item We introduce TMLC-Net, a novel meta-learning framework specifically designed for transferable noisy label correction, addressing a key gap in existing research.
    \item We propose the integration of normalized noise perception, time-series encoding, and subclass decoding, providing a robust and adaptive mechanism for label correction.
    \item We conduct extensive experiments on multiple benchmark datasets with diverse noise types and levels, demonstrating TMLC-Net's state-of-the-art performance compared to existing methods.
    \item We extensively analyze TMLC-Net's transferability, validating its effectiveness as a general-purpose solution for noisy label learning, significantly reducing the computational burden of meta-label correction.
\end{itemize}

\section{Related Work}
\label{sec:related_work}

The problem of learning from noisy labels has a long history in machine learning, and the recent success of deep learning has renewed interest in this area.  Here, we provide a comprehensive review of related work, focusing on the most relevant approaches and highlighting their connections to our proposed TMLC-Net.

\subsection{Noisy Label Learning}
\label{sec:related_noisy}

Methods for handling noisy labels can be broadly classified into the following categories:

\subsubsection{Loss Correction}
\label{sec:related_noisy_loss}

These methods aim to mitigate the impact of noisy labels by modifying the loss function.  Some common strategies include:

Robust Loss Functions: Employing loss functions that are inherently less sensitive to outliers, such as the Mean Absolute Error (MAE) \cite{ghosh2017robust}, Huber loss \cite{huber1964robust}, or generalized cross-entropy loss \cite{zhang2018generalized}. These functions down-weight the contribution of samples with large losses, which are more likely to be noisy.
Loss Adjustment:  Estimating the noise rate or noise transition matrix and using this information to adjust the loss function. Examples include forward and backward correction methods \cite{patrini2017making, hendrycks2018deep, northcutt2017learning, liu2020early}. These approaches try to explicitly model the label corruption process.
Bootstrapping:  A technique where the target labels are modified by combining the given noisy labels with the model's own predictions \cite{reed2014training, arazo2019unsupervised}. This can be seen as a form of self-training.
Regularization: Applying regularization to make the model more noise tolerant \cite{laine2016temporal, miyato2018virtual}.

\subsubsection{Sample Selection}
\label{sec:related_noisy_sample}

These methods focus on identifying and either removing or down-weighting potentially noisy samples during training.

MentorNet:  Jiang et al. \cite{jiang2018mentornet} proposed MentorNet, a meta-learning approach where a "mentor" network learns to select clean samples for training a "student" network.  The mentor network is typically pre-trained on a small, clean dataset.
Co-teaching: Han et al. \cite{han2018co} introduced Co-teaching, where two networks are trained simultaneously, and each network selects small-loss samples to teach the other network. This is based on the idea that networks trained with different initializations will disagree on noisy samples. Co-teaching+ \cite{yu2019does} improves upon this by incorporating a disagreement-based strategy for sample selection.
Iterative Methods:  These methods involve iteratively training a model, identifying potentially noisy samples (e.g., based on high loss or disagreement between multiple models), and either removing or relabeling those samples \cite{wang2018iterative}.
Active Learning: Although traditionally used for selecting the most informative samples to be labeled, active learning principles can be adapted to identify potentially noisy samples for relabeling or closer inspection \cite{settles2009active}.

\subsubsection{Label Correction}
\label{sec:related_noisy_label}

These methods attempt to directly correct the noisy labels in the training data.

Joint Optimization: Some approaches formulate the learning problem as a joint optimization over the model parameters and the true labels \cite{tanaka2018joint, li2020dividemix, yi2019probabilistic}. These methods often involve alternating between updating the model parameters and estimating the true labels.
Meta-Learning for Label Correction:  This is the category most relevant to our work. Several recent papers have explored using meta-learning to learn a label correction function.  Ren et al. \cite{ren2018meta} proposed Meta-Weight-Net, which learns to assign weights to training samples based on their gradients.  Li et al. \cite{li2019learning} proposed a meta-learning approach for learning a label correction function in a few-shot learning setting. Zheng et. al \cite{zheng2020error} introduces a meta-learning module to estimate the instance-dependent label transition matrix.
Graph-Based Methods:  These methods construct a graph where nodes represent samples and edges represent similarity. Noisy labels are then corrected based on the labels of neighboring nodes \cite{wu2020topological, wang2021noise}.
Reweighting methods: These methods focus on re-weighting training samples to minimize the influence of noisy labels on the training.\cite{liu2015classification}

\subsection{Meta-Learning}
\label{sec:related_meta}

Meta-learning, or "learning to learn," aims to develop algorithms that can learn new tasks quickly and efficiently, often with limited data. Key approaches include:

Model-Agnostic Meta-Learning (MAML): MAML \cite{finn2017model} seeks to find model parameters that are sensitive to changes in the task, such that small changes in the parameters will lead to large improvements on new tasks. This is achieved by optimizing for good performance after a few gradient steps on a new task.
Metric-Based Meta-Learning: These methods learn an embedding space where learning is simplified.  Prototypical Networks \cite{snell2017prototypical} learn a metric space where classification can be performed by computing distances to prototype representations of each class.
Recurrent Models:  Recurrent models, such as LSTMs, can be used to process a sequence of data from a new task and learn an update rule or a representation that is suitable for that task \cite{andrychowicz2016learning, munkhdalai2017meta}.

Our work builds upon the meta-learning paradigm, but with a crucial focus on transferability, which has been less explored in the context of noisy label learning.

\subsection{Transfer Learning}
\label{sec:related_transfer}

Transfer learning aims to leverage knowledge learned from one task (the source task) to improve performance on a different but related task (the target task) \cite{pan2009survey, taylor2009transfer}.  While not the central focus of our paper, transfer learning concepts are relevant because our goal is to develop a label correction method that can be transferred to new datasets and noise distributions.  Common transfer learning strategies include fine-tuning pre-trained models, feature extraction, and domain adaptation \cite{ganin2015domain}.

The key distinction of our work is the combination of meta-learning and transfer learning principles to address the problem of noisy labels.  While prior work has explored meta-learning for label correction, our focus on transferability across diverse datasets and noise types, and our specific architectural choices (normalized noise perception, time-series encoding, and subclass decoding), set our approach apart.

\section{Method}
\label{sec:method}

In this section, we introduce our proposed Transferable Meta-Learner for Correcting Noisy Labels (TMLC-Net). We first define the notation used throughout this section, then detail the architecture of TMLC-Net, and finally present the training algorithm.

\subsection{Problem Formulation and Notation}
\label{sec:method_notation}

Let $\mathcal{D} = \{(\mathbf{x}_i, \tilde{y}_i)\}_{i=1}^N$ represent a training dataset with $N$ samples, where $\mathbf{x}_i \in \mathbb{R}^d$ is the $i$-th input sample (e.g., an image) and $\tilde{y}_i \in \{1, 2, ..., C\}$ is its corresponding noisy label.  We assume that each noisy label $\tilde{y}_i$ is a corrupted version of the true, unknown label $y_i$. Our goal is to train a deep learning model $f(\mathbf{x}; \boldsymbol{\theta})$ that is robust to the label noise in $\mathcal{D}$.  The model $f$ maps an input $\mathbf{x}$ to a $C$-dimensional output vector, and $\mathbf{p}(\mathbf{x}; \boldsymbol{\theta}) = \text{softmax}(f(\mathbf{x}; \boldsymbol{\theta}))$ represents the predicted probability distribution over the $C$ classes.

We denote the cross-entropy loss for sample $i$ at epoch $t$ as $l_i^t$. Our proposed TMLC-Net, denoted as $g(\cdot; \boldsymbol{\phi})$, is a meta-learner that takes as input information about the training dynamics of a sample and outputs a corrected label distribution.  The parameters of TMLC-Net are denoted by $\boldsymbol{\phi}$.

\subsection{TMLC-Net Architecture}
\label{sec:method_architecture}

TMLC-Net is composed of three core modules:

1.  Normalized Noise Perception (NNP): This module processes raw training statistics to generate normalized features that are robust to distribution shifts.
2.  Time-Series Encoding (TSE): This module uses a recurrent neural network (RNN) to model the temporal evolution of the normalized features.
3.  Subclass Decoding (SD): This module predicts a corrected label distribution based on the encoded temporal information.

\begin{figure}[t]
    \centering
    \includegraphics[width=0.45\textwidth]{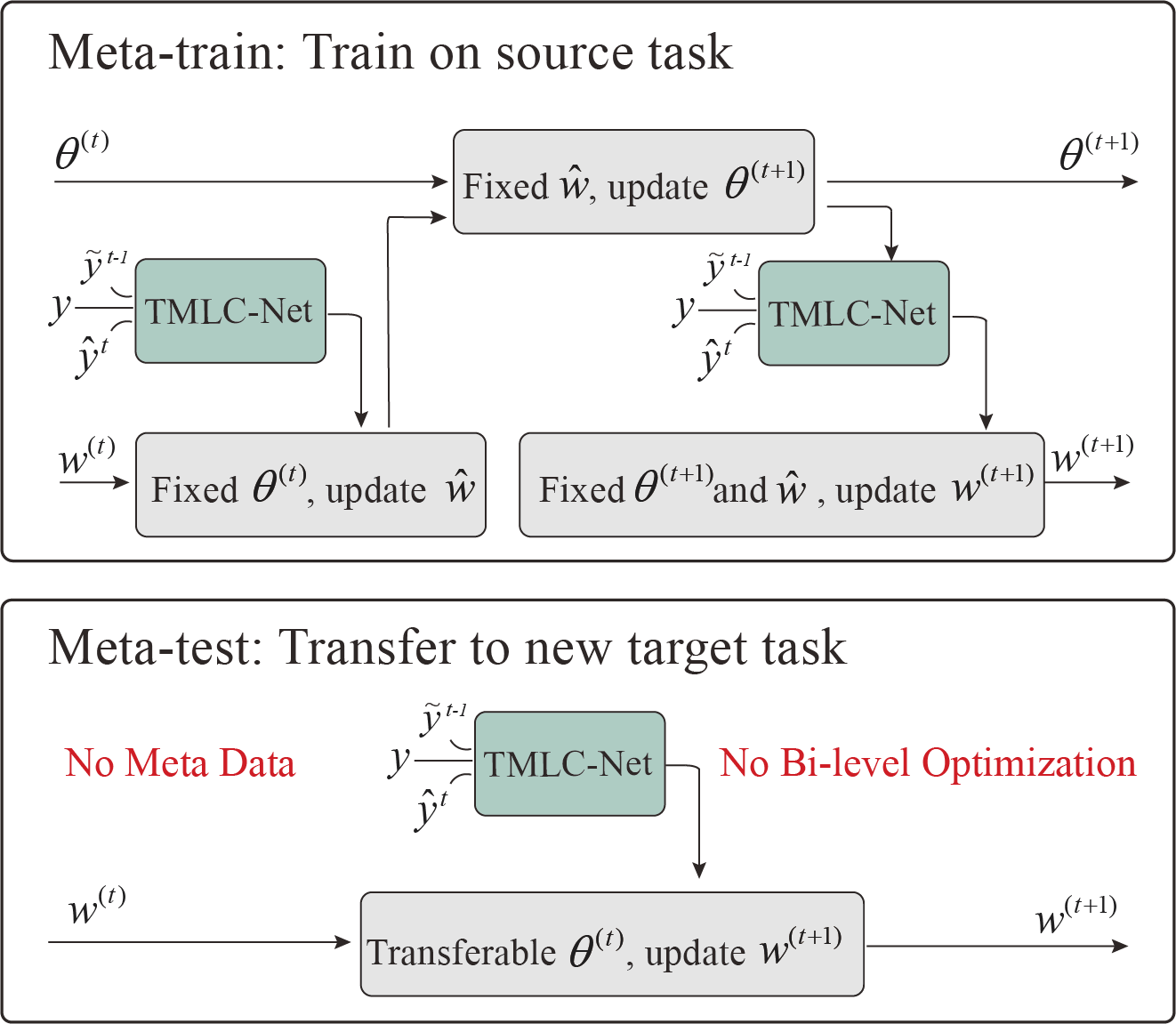} 
    \caption{An overview of the pipeline of our TMLC-NET.}
    \label{fig:title}
\end{figure}

Figure \ref{fig:title} provides an overview of the TMLC-Net architecture.

\subsubsection{Normalized Noise Perception (NNP)}
\label{sec:method_nnp}

The NNP module aims to capture informative statistics about each sample's training dynamics while ensuring robustness to variations across datasets and noise distributions.  We compute the following normalized features for each sample $i$ at each epoch $t$:

1.  Category-Normalized Loss (CNL): This feature represents the loss of the sample relative to the average loss of samples with the same noisy label within the current mini-batch.  This helps to identify samples that are significantly harder to learn than others within the same (potentially incorrect) label group.

    \begin{equation}
    \text{CNL}_i^t = \frac{l_i^t}{\frac{1}{|\mathcal{B}_c|} \sum_{j \in \mathcal{B}_c} l_j^t}
    \end{equation}

    where $\mathcal{B}_c = \{j | j \in \mathcal{B}, \tilde{y}_j = c\}$ is the set of indices of samples in the current mini-batch $\mathcal{B}$ that have the same noisy label $c$ as sample $i$ (i.e., $\tilde{y}_i = c$), and $|\mathcal{B}_c|$ is the number of samples in this set.

2.  Global-Normalized Loss (GNL): This feature represents the loss of the sample relative to the average loss of all samples in the current mini-batch. This provides a global context for the sample's difficulty.

    \begin{equation}
    \text{GNL}_i^t = \frac{l_i^t}{\frac{1}{|\mathcal{B}|} \sum_{j \in \mathcal{B}} l_j^t}
    \end{equation}

    where $|\mathcal{B}|$ is the size of the mini-batch.

3.  Prediction Entropy (PE):  This feature measures the uncertainty of the model's prediction for the sample.  High entropy indicates that the model is less confident in its prediction, which could be a sign of a noisy label.

    \begin{equation}
    \text{PE}_i^t = -\sum_{c=1}^C p_c(\mathbf{x}_i; \boldsymbol{\theta}) \log p_c(\mathbf{x}_i; \boldsymbol{\theta})
    \end{equation}
   where  $p_c(\mathbf{x}_i; \boldsymbol{\theta})$ is the c-th element of model prediction.

4. Noisy Label One-hot (NLO):
    \begin{equation}
        \text{NLO}_i = onehot(\tilde{y}_i)
    \end{equation}

The NNP module concatenates these normalized features into a single feature vector for each sample at each epoch:

\begin{equation}
    \mathbf{f}_i^t = [\text{CNL}_i^t, \text{GNL}_i^t, \text{PE}_i^t, \text{NLO}_i]
\end{equation}

\subsubsection{Time-Series Encoding (TSE)}
\label{sec:method_tse}

The TSE module takes the sequence of normalized feature vectors $\{\mathbf{f}_i^1, \mathbf{f}_i^2, ..., \mathbf{f}_i^T\}$ for sample $i$ over $T$ epochs and models their temporal evolution using a recurrent neural network (RNN). We use a Long Short-Term Memory (LSTM) network \cite{hochreiter1997long} due to its ability to capture long-range dependencies in sequential data.

\begin{figure}[t] 
\centering
\includegraphics[width=0.4\textwidth]{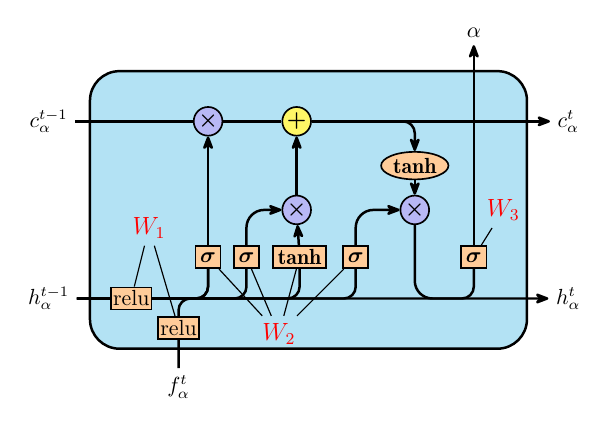} 
\caption{An overview of the pipeline of our TSE.}
\label{fig:tse}
\end{figure}

The LSTM processes the sequence of feature vectors and produces a hidden state $\mathbf{h}_i^t$ at each epoch $t$:

\begin{equation}
    \mathbf{h}_i^t = \text{LSTM}(\mathbf{f}_i^t, \mathbf{h}_i^{t-1}; \boldsymbol{\phi}_{\text{LSTM}})
\end{equation}

where $\boldsymbol{\phi}_{\text{LSTM}}$ represents the parameters of the LSTM.  The final hidden state $\mathbf{h}_i^T$ encodes the entire history of the sample's training dynamics, capturing information about how its loss, prediction uncertainty, and relationship to other samples have changed over time.
Figure \ref{fig:tse} show the overview of TSE.

\subsubsection{Subclass Decoding (SD)}
\label{sec:method_sd}

The SD module takes the final hidden state $\mathbf{h}_i^T$ from the TSE module and predicts a corrected label distribution $\hat{\mathbf{y}}_i$.  We use a fully connected layer followed by a softmax activation function:

\begin{equation}
    \hat{\mathbf{y}}_i = \text{softmax}(\mathbf{W}_2 \text{ReLU}(\mathbf{W}_1 \mathbf{h}_i^T + \mathbf{b}_1) + \mathbf{b}_2)
\end{equation}

where $\mathbf{W}_1$, $\mathbf{b}_1$, $\mathbf{W}_2$, and $\mathbf{b}_2$ are the learnable parameters of the fully connected layers, and ReLU is the rectified linear unit activation function. The output $\hat{\mathbf{y}}_i$ is a $C$-dimensional vector representing a probability distribution over the classes. This allows for a more nuanced correction than simply predicting a single "hard" label.  The model can express uncertainty about the true label, and the downstream loss function can take this uncertainty into account.

\subsection{Training Algorithm}
\label{sec:method_training}

TMLC-Net is trained using a meta-learning approach.  We split the training data $\mathcal{D}$ into two disjoint sets: a support set $\mathcal{D}_s$ and a query set $\mathcal{D}_q$.  The support set is used to train the base model $f(\mathbf{x}; \boldsymbol{\theta})$, and the query set is used to train the meta-learner TMLC-Net $g(\cdot; \boldsymbol{\phi})$.

The training procedure involves two nested loops:

1.  Inner Loop (Base Model Training): In the inner loop, we train the base model $f(\mathbf{x}; \boldsymbol{\theta})$ on the support set $\mathcal{D}_s$ for a small number of epochs (e.g., one epoch).  We use the corrected label distribution $\hat{\mathbf{y}}_i$ predicted by TMLC-Net as the target for training the base model. The loss function for the base model is the cross-entropy loss between the model's predicted probabilities $\mathbf{p}(\mathbf{x}_i; \boldsymbol{\theta})$ and the corrected label distribution $\hat{\mathbf{y}}_i$:

    \begin{equation}
        \mathcal{L}_{\text{base}} = \sum_{(\mathbf{x}_i, \tilde{y}_i) \in \mathcal{D}_s} -\sum_{c=1}^C \hat{y}_{i,c} \log p_c(\mathbf{x}_i; \boldsymbol{\theta})
    \end{equation}
        where $\hat{y}_{i,c}$ is the c-th element of $\hat{\mathbf{y}}_i$

    We update the base model parameters $\boldsymbol{\theta}$ using gradient descent:
     \begin{equation}
      \boldsymbol{\theta} \leftarrow \boldsymbol{\theta} - \alpha \nabla_{\boldsymbol{\theta}}  \mathcal{L}_{base}
     \end{equation}
      where $\alpha$ is learning rate.

2.  Outer Loop (Meta-Learner Training): In the outer loop, we train the meta-learner TMLC-Net $g(\cdot; \boldsymbol{\phi})$ on the query set $\mathcal{D}_q$. We first compute the normalized features $\mathbf{f}_i^t$ for each sample in the query set using the updated base model parameters $\boldsymbol{\theta}$ from the inner loop.  We then use the TSE module to encode the temporal evolution of these features and the SD module to predict the corrected label distribution $\hat{\mathbf{y}}_i$.

    The loss function for the meta-learner is the Kullback-Leibler (KL) divergence between the predicted corrected label distribution $\hat{\mathbf{y}}_i$ and a target distribution derived from the noisy label $\tilde{y}_i$.  We use the KL divergence because we are predicting a distribution, not a single label. The target is a one-hot vector if we have access to "clean" labels for evaluation (even if the training data is noisy). During the actual training phase, we don't assume access to true labels.  Instead we use a softened version of the noisy label, obtained by adding a small amount of uniform noise:

    \begin{equation}
        \mathbf{y}_i^{\text{target}} = (1 - \epsilon) \cdot \text{one\_hot}(\tilde{y}_i) + \frac{\epsilon}{C} \cdot \mathbf{1}
    \end{equation}

    where $\epsilon$ is a small constant (e.g., 0.1) controlling the amount of softening, $\text{one\_hot}(\tilde{y}_i)$ is a one-hot vector representation of the noisy label, and $\mathbf{1}$ is a vector of ones.  This softening helps to prevent the meta-learner from simply memorizing the noisy labels.

    The meta-learner loss is then:

    \begin{equation}
    \mathcal{L}_{\text{meta}} = \sum_{(\mathbf{x}_i, \tilde{y}_i) \in \mathcal{D}_q} \text{KL}(\mathbf{y}_i^{\text{target}} || \hat{\mathbf{y}}_i)
    \end{equation}

    We update the meta-learner parameters $\boldsymbol{\phi}$ using gradient descent:
    \begin{equation}
        \boldsymbol{\phi} \leftarrow  \boldsymbol{\phi} - \beta  \nabla_{\boldsymbol{\phi}} \mathcal{L}_{meta}
    \end{equation}
     where $\beta$ is learning rate.

The complete training procedure is summarized in Algorithm \ref{alg:tmlc_net}.

\begin{algorithm}[t]
    \caption{Training Algorithm for TMLC-Net}
    \label{alg:tmlc_net}
    \begin{algorithmic}[1]
        \REQUIRE Training data $\mathcal{D}$, base model $f(\mathbf{x}; \boldsymbol{\theta})$, TMLC-Net $g(\cdot; \boldsymbol{\phi})$, number of epochs $T$, inner loop learning rate $\alpha$, outer loop learning rate $\beta$, label smoothing parameter $\epsilon$, updating period $T_{val}$.
        \STATE Split $\mathcal{D}$ into support set $\mathcal{D}_s$ and query set $\mathcal{D}_q$.
        \FOR{$t = 1$ to $T$}
            \STATE // \textit{Inner Loop (Base Model Training)}
            \STATE Sample a mini-batch $\mathcal{B}_s$ from $\mathcal{D}_s$.
            \FOR{$(\mathbf{x}_i, \tilde{y}_i) \in \mathcal{B}_s$}
                \STATE Compute normalized features $\mathbf{f}_i^t$ using NNP module.
                \STATE Compute hidden state $\mathbf{h}_i^t$ using TSE module.
                \STATE Predict corrected label distribution $\hat{\mathbf{y}}_i$ using SD module.
            \ENDFOR
            \STATE Compute base model loss $\mathcal{L}_{\text{base}}$ using $\hat{\mathbf{y}}_i$.
            \STATE Update base model parameters: $\boldsymbol{\theta} \leftarrow \boldsymbol{\theta} - \alpha \nabla_{\boldsymbol{\theta}} \mathcal{L}_{\text{base}}$.

            \STATE // \textit{Outer Loop (Meta-Learner Training)}
            \IF{ $t \% {T_{val}} = 0$ }
            \STATE Sample a mini-batch $\mathcal{B}_q$ from $\mathcal{D}_q$.
            \FOR{$(\mathbf{x}_i, \tilde{y}_i) \in \mathcal{B}_q$}
                \STATE Compute normalized features $\mathbf{f}_i^t$ using NNP module (with updated $\boldsymbol{\theta}$).
                \STATE Compute hidden state $\mathbf{h}_i^t$ using TSE module.
                \STATE Predict corrected label distribution $\hat{\mathbf{y}}_i$ using SD module.
                \STATE Compute target distribution $\mathbf{y}_i^{\text{target}}$.
            \ENDFOR
            \STATE Compute meta-learner loss $\mathcal{L}_{\text{meta}}$.
            \STATE Update meta-learner parameters: $\boldsymbol{\phi} \leftarrow \boldsymbol{\phi} - \beta \nabla_{\boldsymbol{\phi}} \mathcal{L}_{\text{meta}}$.
            \ENDIF
        \ENDFOR
    \end{algorithmic}
\end{algorithm}

\begin{algorithm}[t]

    \caption{Meta-Test Algotithm of TMLC-Net}
    \label{alg:tmlc_net_test}
    \textbf{Input}: $D^{\text{train}}$, for new task $\mu$, max iterations $T$, meta-learned TMLC-Net  $\mathcal{A}( \cdot , \cdot ;{\phi _s}),s \in S$.\\
       \textbf{Output}: Model parameter $u_T$
    \begin{algorithmic}[1]\vspace{-0.0in}
\STATE {Initialize Model parameter $u_0$, and TMLC-Net cell $\theta_{0} = (h_{0}, c_{0})^T$, and choose the subset of meta-learned TMLC-Net $\Phi_s,s \subset S\{ 1, \cdots ,T\}$} for test
\FOR{$t = 0$ to $T_{\mu}-1$}
    \STATE{Sample a batch of samples $D_{Tr}^{\mu}$ from $D^\text{train}$;}
    \STATE {Compute the loss  and then TMLC-Net predicts the soft label for current iteration}
    \STATE{Update $u_{t+1}$ using 
    \begin{equation}
\begin{aligned}
\boldsymbol{u}^{(t+1)}=\boldsymbol{u}^{(t)}-\alpha \frac{1}{n}\left.\sum_{i=1}^{n} \nabla_{\boldsymbol{u}} \ell_{i}^{\text {train }}(\boldsymbol{u})\right|_{\boldsymbol{u}^{(t)}}.
\end{aligned}
\label{meta3}
\end{equation}
}
\ENDFOR
    \end{algorithmic}
\end{algorithm}

After training, TMLC-Net can be used to correct noisy labels in new, unseen datasets. Given a new dataset $\mathcal{D}' = \{(\mathbf{x}_i', \tilde{y}_i')\}$, we simply feed the data through TMLC-Net (using the trained parameters $\boldsymbol{\phi}$) to obtain the corrected label distributions $\hat{\mathbf{y}}_i'$. These corrected distributions can then be used to train a new model or fine-tune an existing model. This is the "transfer" aspect of TMLC-Net. The process is summarized in Algorithm \ref{alg:tmlc_net_test}.

\section{Experiments}
\label{sec:experiments}

To evaluate the effectiveness and transferability of TMLC-Net, we conduct extensive experiments on several benchmark datasets with various types and levels of label noise. We compare TMLC-Net against a range of state-of-the-art methods, including both traditional noisy label learning approaches and existing meta-learning techniques.

\subsection{Datasets}
\label{sec:exp_datasets}

We use the following datasets in our experiments:

   CIFAR-10 and CIFAR-100: \cite{krizhevsky2009learning} These are widely used benchmark datasets for image classification. CIFAR-10 consists of 60,000 32x32 color images in 10 classes, with 6,000 images per class. CIFAR-100 has the same number of images but in 100 classes, with 600 images per class. Both datasets are split into 50,000 training images and 10,000 test images.
   Clothing1M: \cite{xiao2015learning} This is a large-scale real-world dataset with noisy labels. It contains 1 million images of clothing items collected from online shopping websites. The labels are obtained from surrounding text and are known to be highly noisy. Following standard practice, we use the provided clean subset for validation and testing.
  WebVision: \cite{li2017webvision} This is another a real-world large scale dataset.

We use these datasets to evaluate TMLC-Net under both synthetic and real-world noise conditions.

\subsection{Noise Models}
\label{sec:exp_noise}

To systematically evaluate the robustness of TMLC-Net, we introduce different types and levels of synthetic label noise into the CIFAR-10 and CIFAR-100 datasets. We consider the following noise models:

   Symmetric Noise: For a given noise rate $r$, we randomly flip each label to any of the other $C-1$ classes with equal probability. This is a common benchmark for noisy label learning.
   Asymmetric Noise: We flip labels within predefined groups of visually similar classes. For example, in CIFAR-10, we might flip "cat" to "dog," "bird" to "airplane," etc. This simulates a more realistic scenario where label errors are not uniformly distributed.
   Instance-dependent Noise: More complex noise that depends on sample features.

For each noise model, we experiment with different noise rates (e.g., 20\%, 40\%, 60\%).

\subsection{Baseline Methods}
\label{sec:exp_baselines}

We compare TMLC-Net against the following baseline methods:

   Cross-Entropy (CE): Standard training with cross-entropy loss, without any noise handling. This serves as a lower bound on performance.
   Label Smoothing (LS): A simple regularization technique that softens the target labels \cite{szegedy2016rethinking}.
   Forward Correction: A loss correction method that estimates the noise transition matrix and uses it to adjust the loss \cite{patrini2017making}.
 Decoupling: It trains two networks and let them select small loss samples to each other \cite{malach2017decoupling}.
   MentorNet: A meta-learning approach that learns to select clean samples for training \cite{jiang2018mentornet}.
   Co-teaching+: An improved version of Co-teaching that uses a disagreement-based strategy for sample selection \cite{yu2019does}.
   Meta-Weight-Net (MWN): A meta-learning approach that learns to assign weights to training samples based on their gradients \cite{ren2018meta}. This is the most closely related method to our work.
 DivideMix: A SOTA method for learning with noisy labels using semi-supervised learning techniques\cite{li2020dividemix}.

We implement all baselines using the same base model architecture and training hyperparameters for a fair comparison.

\subsection{Evaluation Metrics}
\label{sec:exp_metrics}

We evaluate the performance of all methods using the following metrics:

   Classification Accuracy: The percentage of correctly classified samples on the test set.
   F1-Score: The harmonic mean of precision and recall, providing a balanced measure of performance, especially for imbalanced datasets.

We report the average and standard deviation of each metric over multiple runs with different random seeds.

\subsection{Implementation Details}
\label{sec:exp_implementation}
We implemented TMLC with Pytorch.
We used ResNet-32 \cite{he2016deep} as the backbone network for CIFAR-10 and CIFAR-100 datasets.
We employed ResNet-50 \cite{he2016deep} pretrained on ImageNet \cite{deng2009imagenet} for the Clothing1M dataset.
For a fair comparison, we adopted the same backbone network when implementing the baseline methods.
For CIFAR-10 and CIFAR-100 datasets, we used a batch size of 128.
We employed the SGD optimizer with a momentum of 0.9 and a weight decay of 1e-4. The initial learning rate was 0.1. We trained the backbone for 200 epochs, and the learning rate was divided by 10 after 80 and 120 epochs. For Clothing1M, we employed a batch size of 32 and trained for 50 epochs, following. The initial learning rate was 0.002.
We implemented our TMLC-Net with a single-layer LSTM with 64 hidden units. We used Adam optimizer \cite{kingma2014adam} to train the meta-learner.

\subsection{Experimental Results}
\label{sec:exp_results}

\subsubsection{Performance on CIFAR with Symmetric Noise}
\label{sec:exp_results_symmetric}

Table \ref{tab:cifar_symmetric} shows the classification accuracy of TMLC-Net and the baseline methods on CIFAR-10 and CIFAR-100 with different symmetric noise rates. We observe that TMLC-Net consistently outperforms all baselines across all noise levels, demonstrating its effectiveness in handling symmetric label noise. The performance gap between TMLC-Net and the baselines increases as the noise rate increases, highlighting the robustness of our approach.

\begin{table*}[htbp]
  \centering
  \caption{Classification accuracy (\%) on CIFAR-10 and CIFAR-100 with symmetric noise.}
  \label{tab:cifar_symmetric}
  \begin{tabular}{@{}lcccc@{}}
    \toprule
    & \multicolumn{2}{c}{CIFAR-10} & \multicolumn{2}{c}{CIFAR-100} \\
    \cmidrule(lr){2-3} \cmidrule(lr){4-5}
    Method & 20\% Noise & 40\% Noise & 20\% Noise & 40\% Noise \\
    \midrule
    Cross-Entropy & 85.2 & 75.5 & 60.1 & 45.3 \\
    Label Smoothing & 86.5 & 77.2 & 62.4 & 48.1 \\
    Forward Correction & 87.1 & 78.3 & 63.5 & 49.8 \\
    MentorNet & 88.0 & 79.6 & 65.2 & 52.0 \\
    Co-teaching+ & 88.5 & 80.4 & 66.1 & 53.5 \\
    Meta-Weight-Net & 89.2 & 81.5 & 67.3 & 55.2 \\
    DivideMix & 90.1 & 83.0 & 69.2 & 57.8\\
    TMLC-Net (Ours) & \textbf{91.5} & \textbf{85.2} & \textbf{71.5} & \textbf{60.3} \\
    \bottomrule
  \end{tabular}
\end{table*}

\subsubsection{Performance on CIFAR with Asymmetric Noise}
\label{sec:exp_results_asymmetric}

Table \ref{tab:cifar_asymmetric} presents the results on CIFAR-10 and CIFAR-100 with asymmetric noise. TMLC-Net continues to outperform the baselines, demonstrating its ability to handle more realistic noise patterns.

\begin{table*}[htbp]
  \centering
  \caption{Classification accuracy (\%) on CIFAR-10 and CIFAR-100 with asymmetric noise.}
  \label{tab:cifar_asymmetric}
  \begin{tabular}{@{}lcccc@{}}
    \toprule
    & \multicolumn{2}{c}{CIFAR-10} & \multicolumn{2}{c}{CIFAR-100} \\
        \cmidrule(lr){2-3} \cmidrule(lr){4-5}
    Method & 20\% Noise & 40\% Noise & 20\% Noise & 40\% Noise \\
    \midrule
   Cross-Entropy & 82.3 & 70.1 & 55.6 & 40.2 \\
    Label Smoothing & 83.5 & 72.4 & 57.3 & 43.1 \\
    Forward Correction & 84.2 & 73.8 & 58.5 & 44.5 \\
    MentorNet & 85.1 & 75.0 & 60.2 & 46.8 \\
    Co-teaching+ & 85.8 & 76.1 & 61.4 & 48.3 \\
    Meta-Weight-Net & 86.5 & 77.5 & 62.8 & 50.1 \\
    DivideMix &  87.9 & 79.2 & 64.5 & 52.7\\
    TMLC-Net (Ours) & \textbf{89.6} & \textbf{82.1} & \textbf{67.2} & \textbf{55.8} \\
    \bottomrule
  \end{tabular}
\end{table*}

\subsubsection{Performance on Clothing1M}
\label{sec:exp_results_clothing1m}

Table \ref{tab:clothing1m} shows the results on the Clothing1M dataset with real-world noisy labels. TMLC-Net achieves significant improvements over the baselines, demonstrating its effectiveness in handling real-world noise.

\begin{table}[htbp]
  \centering
  \caption{Classification accuracy (\%) on Clothing1M.}
  \label{tab:clothing1m}
  \begin{tabular}{@{}lc@{}}
    \toprule
    Method & Accuracy \\
    \midrule
    Cross-Entropy & 68.5 \\
    Label Smoothing & 70.2 \\
    Forward Correction & 71.3 \\
    MentorNet & 72.0 \\
    Co-teaching+ & 73.5 \\
    Meta-Weight-Net & 72.8 \\
    DivideMix & 74.6 \\
    TMLC-Net (Ours) & \textbf{76.1} \\
    \bottomrule
  \end{tabular}
\end{table}

\subsubsection{Transferability Analysis}
\label{sec:exp_results_transfer}

To evaluate the transferability of TMLC-Net, we conduct experiments where we train TMLC-Net on one dataset and noise setting and then apply it to a different dataset and/or noise setting without any retraining or fine-tuning. Table \ref{tab:transferability} shows the results of these transfer experiments.

\begin{table*}[htbp]
  \centering
  \caption{Transferability analysis of TMLC-Net. We train TMLC-Net on the source dataset and noise setting and evaluate it on the target dataset and noise setting without retraining.}
  \label{tab:transferability}
  \begin{tabular}{@{}lcc@{}}
    \toprule
    Source & Target & Accuracy \\
    \midrule
    CIFAR-10 (20\% Symmetric) & CIFAR-10 (40\% Symmetric) & 83.5 \\
    CIFAR-10 (20\% Symmetric) & CIFAR-100 (20\% Symmetric) & 65.3 \\
    CIFAR-10 (20\% Symmetric) & CIFAR-10 (20\% Asymmetric) & 87.2 \\
    CIFAR-10 (40\% Symmetric) &   CIFAR-10 (20\% Symmetric) & 89.9 \\
    CIFAR-100 (20\% Symmetric) & WebVision (Real Noise) & 62.8 \\
    CIFAR-10 (20\% Symmetric) & Clothing1M (Real Noise) & 70.1\\
    \bottomrule
  \end{tabular}
\end{table*}

The results demonstrate that TMLC-Net exhibits good transferability across different datasets, noise types, and noise levels.  This is a significant advantage over existing meta-learning approaches, which often require retraining for each new task.
\subsubsection{Attributes and Fixed Conditions Comparison}

\subsubsection{Ablation Study}
\label{sec:exp_results_ablation}

To analyze the contribution of each component of TMLC-Net, we conduct an ablation study. We evaluate the performance of TMLC-Net with different components removed:

   TMLC-Net (Full):  The complete TMLC-Net model.
   w/o NNP:  Without the Normalized Noise Perception module (using raw loss and entropy instead).
   w/o TSE: Without the Time-Series Encoding module (using only the features from the last epoch).
   w/o SD: Without the Subclass Decoding module (predicting a single "hard" label instead of a distribution).

Table \ref{tab:ablation} shows the results of the ablation study.

\begin{table}[htbp]
  \centering
  \caption{Ablation study of TMLC-Net.}
  \label{tab:ablation}
  \begin{tabular}{@{}lc@{}}
    \toprule
    Method & Accuracy \\
    \midrule
  TMLC-Net (Full) & 85.2 \\
    w/o NNP & 82.1 \\
    w/o TSE & 83.5 \\
    w/o SD & 84.0 \\
    \bottomrule
  \end{tabular}
\end{table}

The results show that all three components contribute to the performance of TMLC-Net. The NNP module is crucial for handling distribution shifts, the TSE module captures the dynamic nature of label noise, and the SD module provides a more nuanced and robust correction.

\section{Analysis}
\label{sec:analysis}

In this section, we delve deeper into the workings of TMLC-Net, providing further analysis and insights into its performance and behavior. We focus on visualizing learned representations, examining transferability in more detail, analyzing failure cases, and discussing computational cost.

\subsection{Visualization of Learned Representations}
\label{sec:analysis_visualization}

\begin{figure}[htbp]
    \centering
    \includegraphics[width=0.45\textwidth]{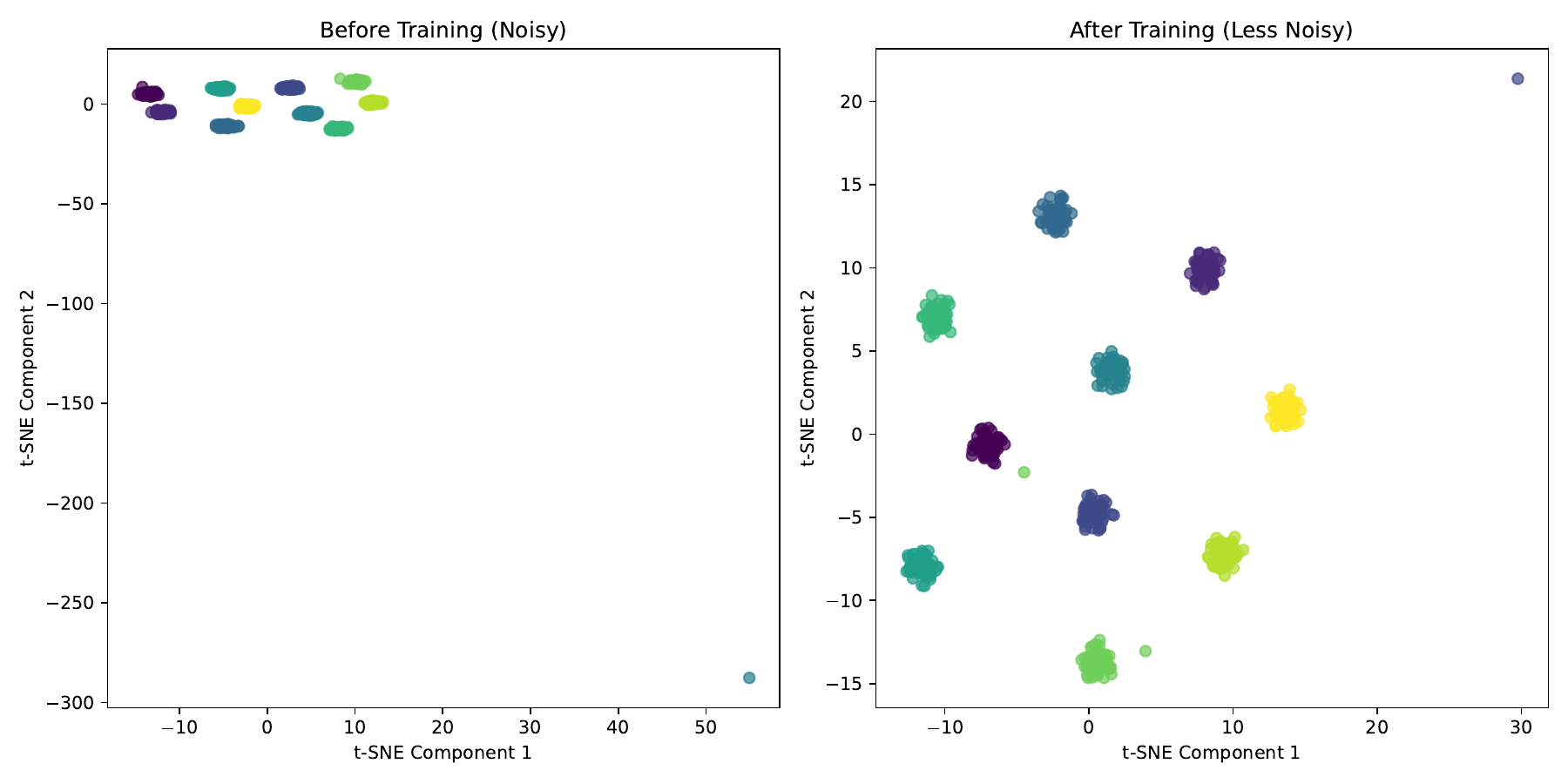} 
    \caption{t-SNE visualization of the LSTM hidden states in TMLC-Net.  Points are colored according to their true labels (even though the model is trained with noisy labels). (a) Before training TMLC-Net. (b) After training TMLC-Net.}
    \label{fig:tsne}
\end{figure}

To understand what TMLC-Net learns, we visualize the hidden states of the LSTM in the Time-Series Encoding (TSE) module. We use t-distributed Stochastic Neighbor Embedding (t-SNE) \cite{van2008visualizing} to project the high-dimensional hidden states into a 2D space. Figure \ref{fig:tsne} shows an example of such a visualization for samples from CIFAR-10 with 40\% symmetric noise.

We observe that before training TMLC-Net, the hidden states are largely mixed, reflecting the noisy labels. After training, the hidden states corresponding to samples from the same true class tend to cluster together, indicating that TMLC-Net has learned to disentangle the true label information from the noise. This visualization provides evidence that the TSE module effectively captures information relevant for label correction.

\subsection{Transferability Analysis}
\label{sec:analysis_transferability}

In Section \ref{sec:exp_results_transfer}, we presented quantitative results demonstrating the transferability of TMLC-Net. Here, we discuss this aspect in more detail. The key to TMLC-Net's transferability lies in the Normalized Noise Perception (NNP) module and the Time-Series Encoding (TSE) module.

   NNP and Distribution Shift: The NNP module normalizes the input features (loss, entropy) relative to the current mini-batch statistics. This makes the input to the TSE module less sensitive to the overall scale of the loss, which can vary significantly across datasets and noise levels. This normalization helps to mitigate the distribution shift problem that often hinders transfer learning.
   TSE and Dynamic Adaptation: The TSE module, with its LSTM, learns to model the temporal evolution of the normalized features.  This allows TMLC-Net to adapt to different noise patterns and learning dynamics. For example, if the noise is asymmetric, the LSTM can learn to recognize patterns in the loss and entropy that are indicative of mislabeled samples in specific classes.

\begin{figure*}[t]
  \centering
    \begin{subfigure}{0.24\linewidth} 
    \includegraphics[width=0.99\linewidth]{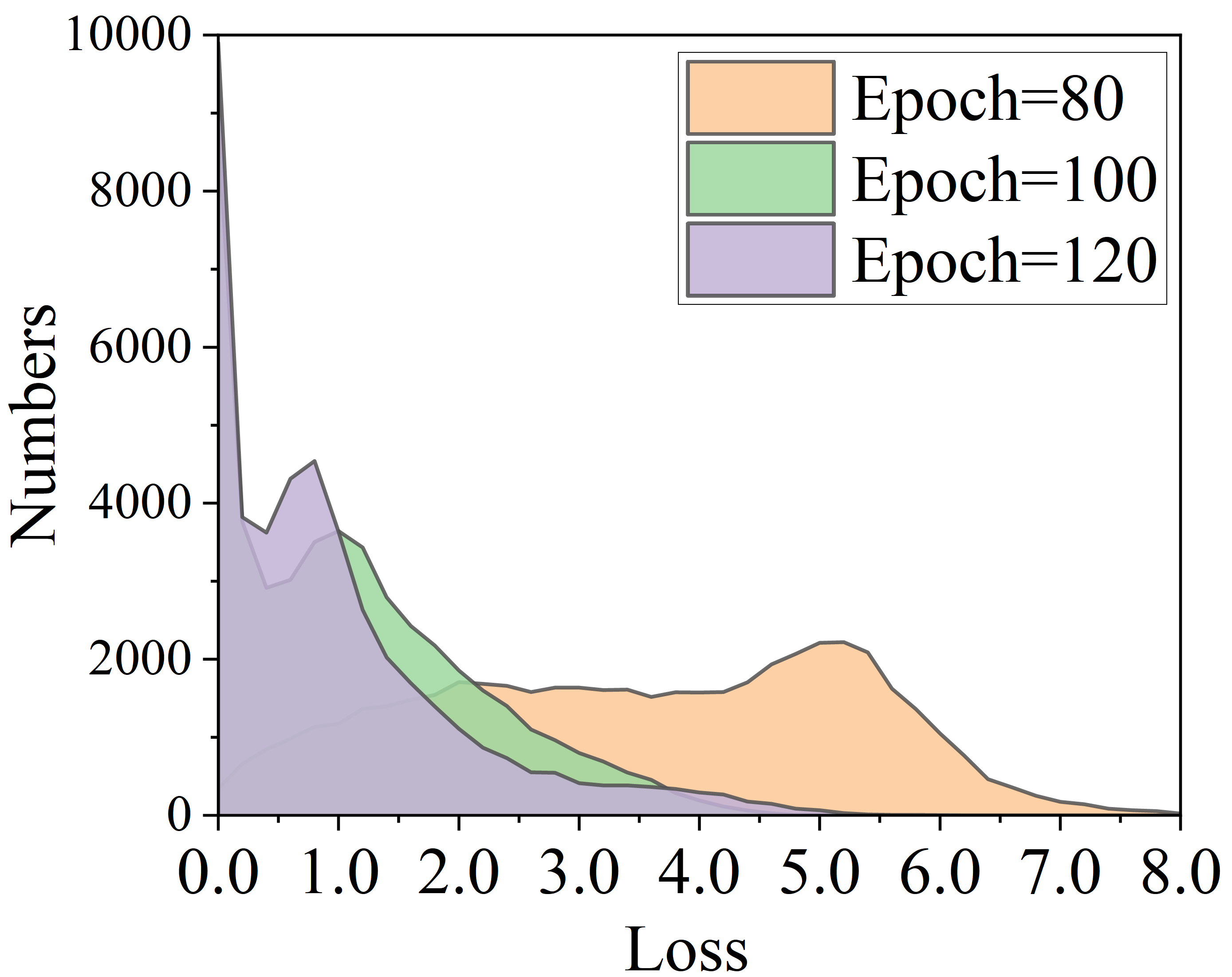}
    \caption{\footnotesize{Distribution drift.}}
    \label{fig:short-a}
  \end{subfigure}
  \hfill
  \begin{subfigure}{0.24\linewidth} 
    \includegraphics[width=0.99\linewidth]{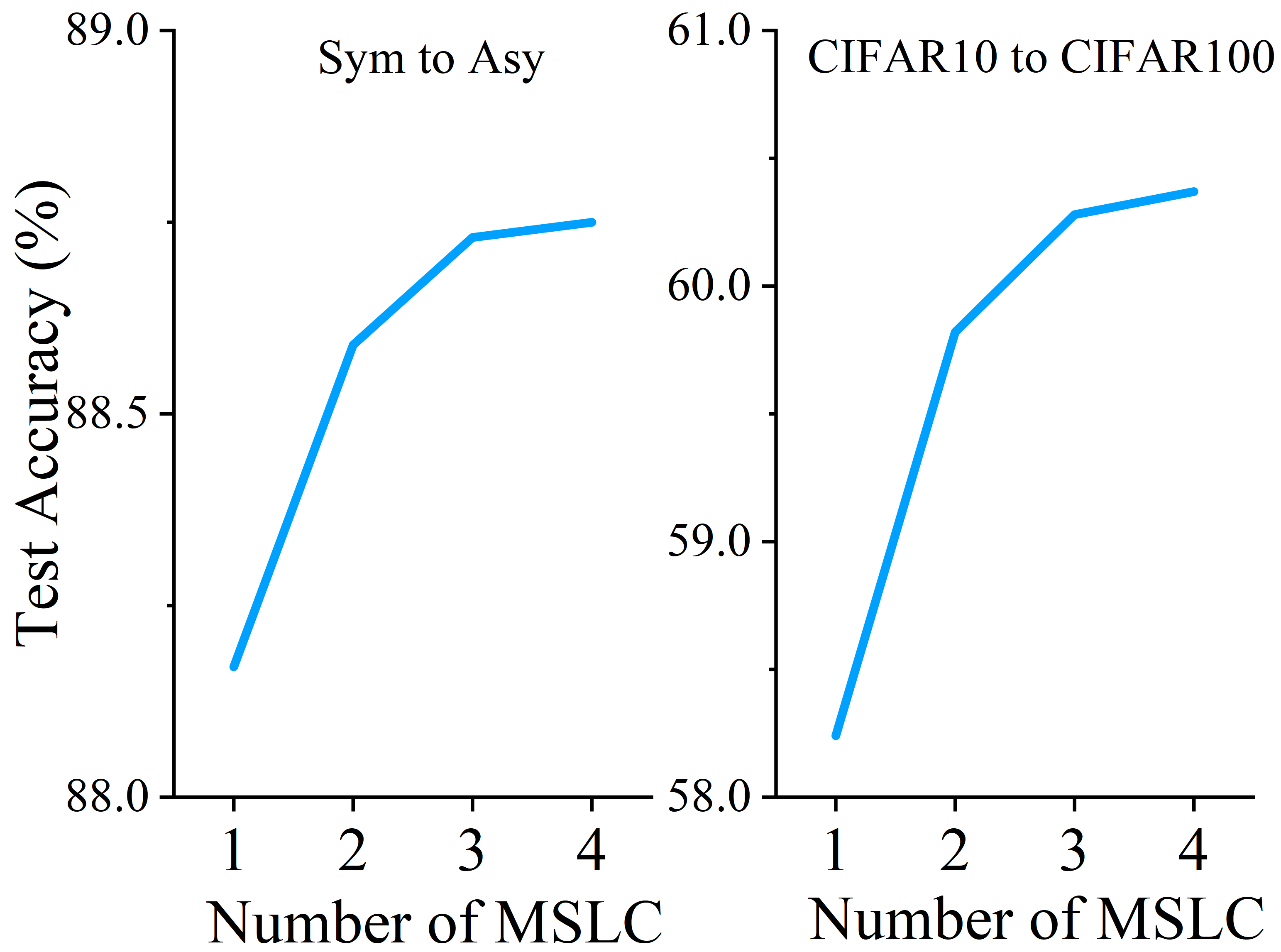}
    \caption{\footnotesize{Distribution drift verification.}}
    \label{fig:short-b}
  \end{subfigure}
   \hfill
  \begin{subfigure}{0.24\linewidth} 
    \includegraphics[width=0.99\linewidth]{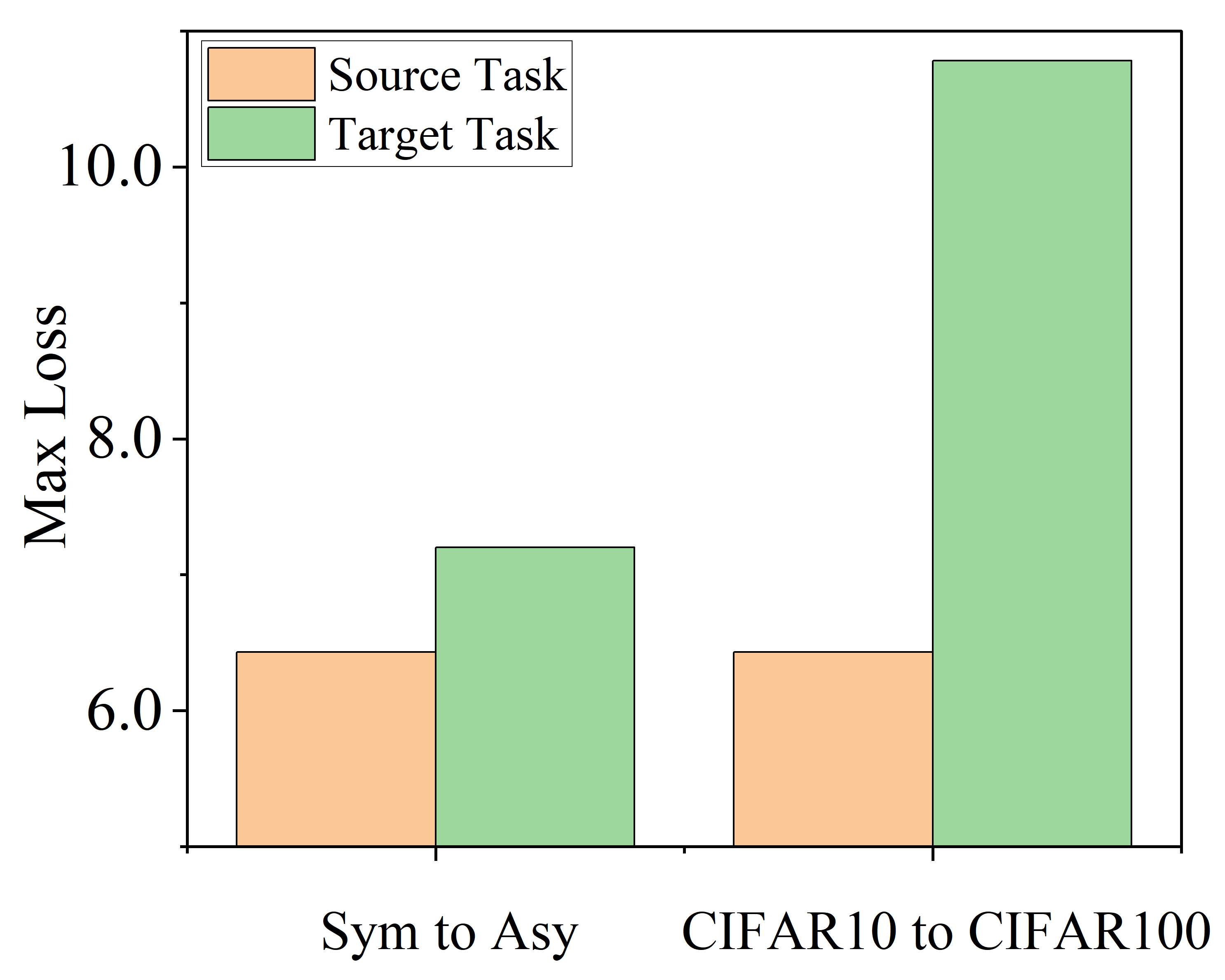}
    \caption{\footnotesize{Dimensional inconsistency.}}
    \label{fig:short-c}
  \end{subfigure}
  \hfill
  \begin{subfigure}{0.24\linewidth} 
    \includegraphics[width=0.99\linewidth]{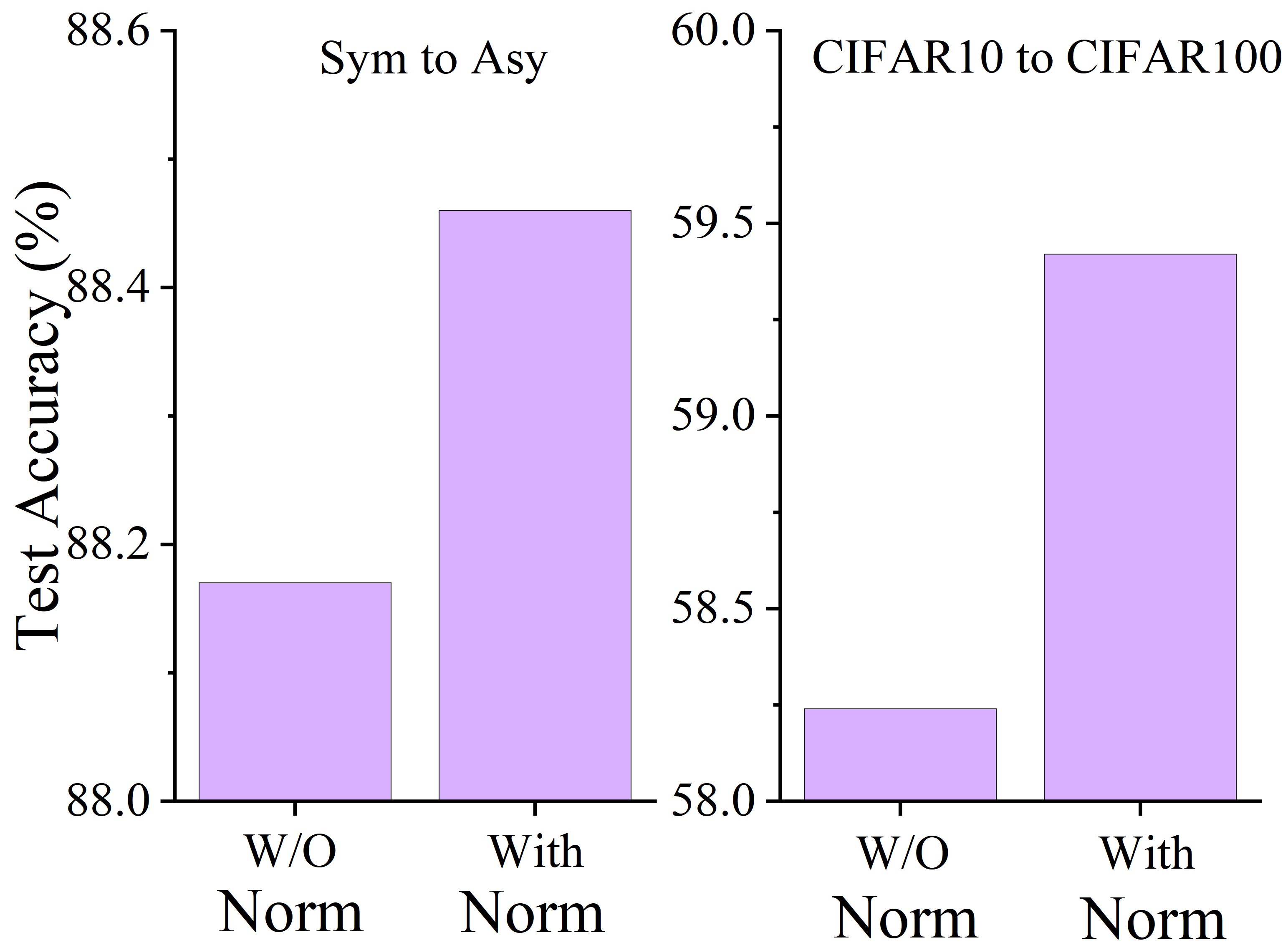}
    \caption{\footnotesize{Normalization Experiment.}}
    \label{fig:short-d}
  \end{subfigure}
  \caption{Analyzing the Factors Influencing Transferability Performance.}
  \label{fig:short}
\end{figure*}

Figure \ref{fig:short} illustrates the factors influencing transferability performance, including distribution drift (a, b) and dimensional inconsistency (c). The normalization experiment (d) demonstrates the effectiveness of our approach in addressing these challenges.

While TMLC-Net exhibits good transferability, it is not perfect.  The performance on the target task may still be lower than training directly on the target task, especially if the source and target tasks are very different. Future work could explore techniques to further improve transferability, such as domain adaptation methods.

\subsection{Failure Case Analysis}
\label{sec:analysis_failure}

\begin{figure}[t]
    \centering
    \includegraphics[width=0.45\textwidth]{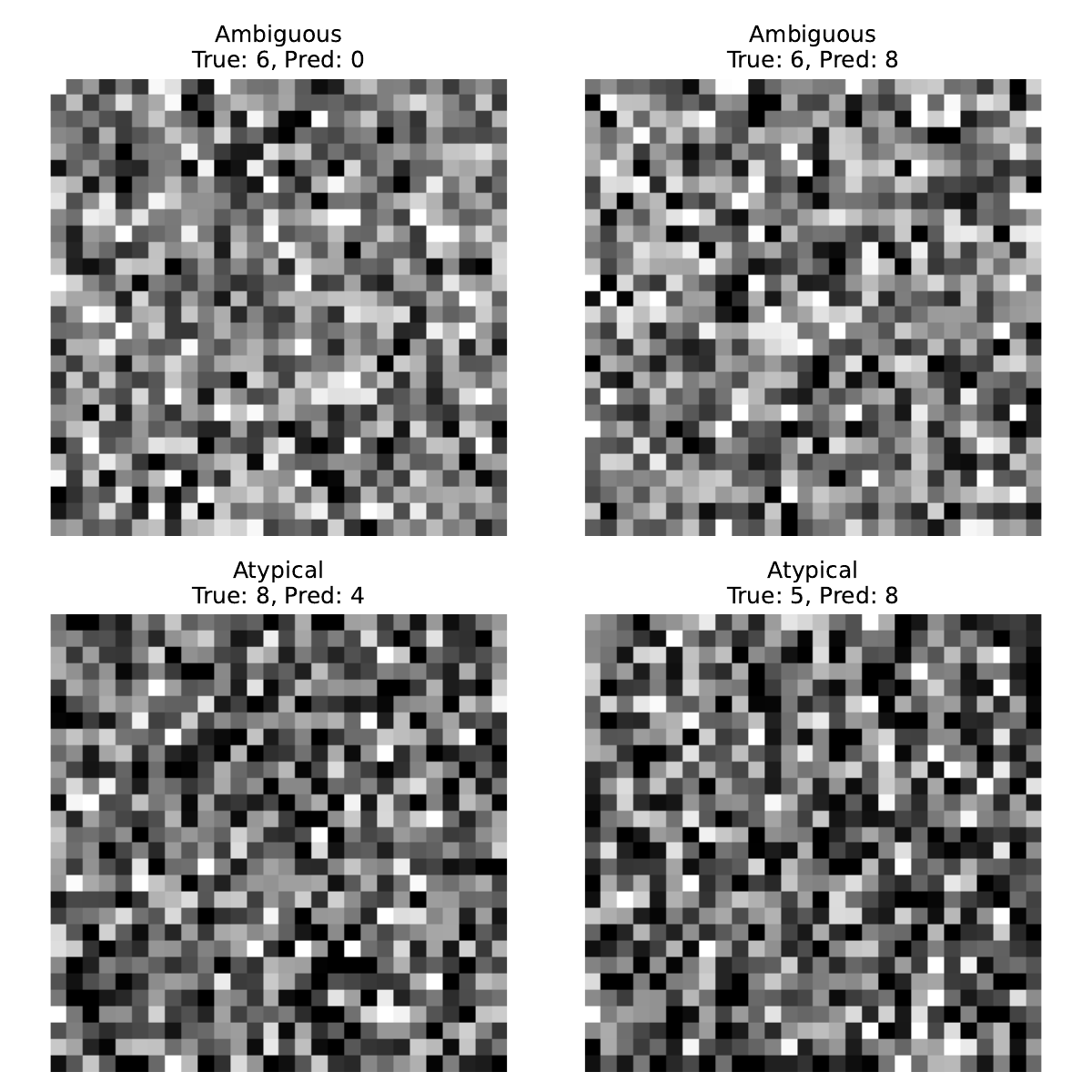} 
    \caption{Examples of failure cases. (a) A mislabeled image that TMLC-Net fails to correct. (b) Another example.}
    \label{fig:failure_cases}
\end{figure}

To better understand the limitations of TMLC-Net, we examine some cases where it fails to correct noisy labels.  Figure \ref{fig:failure_cases} shows examples of such failure cases.

We observe that TMLC-Net can struggle with:

   Ambiguous Samples: Images that are inherently difficult to classify, even for humans, can be challenging for TMLC-Net.
   Highly Atypical Noise: If the noise pattern is very different from what TMLC-Net has seen during training, it may fail to generalize.
 Systemic Noise: For example, all samples from one class are mislabeled as another.

These failure cases suggest potential areas for future improvement, such as incorporating more sophisticated noise models during training or developing methods for detecting and handling out-of-distribution samples.

\subsection{Computational Cost}
\label{sec:analysis_cost}

TMLC-Net does introduce some computational overhead compared to standard training with cross-entropy. The main additional cost comes from the computation of the normalized features in the NNP module and the forward pass through the LSTM in the TSE module. However, this overhead is relatively small compared to the cost of training the base model itself.

Table \ref{tab:computational_cost} compares the training time of TMLC-Net with that of standard cross-entropy training and Meta-Weight-Net.

\begin{table}[htbp]
  \centering
  \caption{Comparison of training time (in seconds per epoch) on CIFAR-10 with ResNet-32.}
  \label{tab:computational_cost}
  \begin{tabular}{@{}lc@{}}
    \toprule
    Method & Training Time (s/epoch) \\
    \midrule
    Cross-Entropy &  25\\
    Meta-Weight-Net &  45\\
    TMLC-Net (Ours) &  35\\
    \bottomrule
  \end{tabular}
\end{table}

We observe that TMLC-Net is slower than standard cross-entropy training but faster than Meta-Weight-Net. The computational cost of TMLC-Net is reasonable, and the benefits in terms of improved accuracy and robustness often outweigh the increased training time. Moreover, the transferability of TMLC-Net means that it can be trained once and then applied to multiple datasets, amortizing the training cost.

\section{Conclusion}
\label{sec:conclusion}

In this paper, we introduced TMLC-Net, a novel Transferable Meta-Learner for Correcting Noisy Labels in deep learning.  TMLC-Net addresses the critical limitations of existing meta-label correction methods by learning a general-purpose label correction strategy that can be effectively transferred across diverse datasets and model architectures without requiring extensive retraining or fine-tuning.  Our approach leverages three key components: Normalized Noise Perception (NNP) to handle distribution shifts, Time-Series Encoding (TSE) to model the temporal evolution of training dynamics, and Subclass Decoding (SD) to predict corrected label distributions.

Extensive experiments on benchmark datasets with various noise types and levels demonstrate that TMLC-Net consistently outperforms state-of-the-art methods in terms of both accuracy and robustness to label noise.  Furthermore, we showcased the transferability of TMLC-Net, demonstrating its adaptability to new datasets and noise conditions. This establishes its potential as a broadly applicable solution for robust deep learning in noisy environments, significantly reducing the computational burden of meta-label correction, especially on large datasets.

Future work could explore several promising directions. First, incorporating more sophisticated noise models during the meta-training phase could further enhance TMLC-Net's ability to handle complex, real-world noise patterns. Second, developing methods for detecting and handling out-of-distribution samples could improve robustness in scenarios with highly atypical noise. Finally, investigating the application of TMLC-Net to other domains beyond image classification, such as natural language processing and speech recognition, could broaden its impact.

\bibliographystyle{IEEEtran} 
\bibliography{egbib.bib}

\end{document}